\documentclass[conference]{IEEEtran}
\IEEEoverridecommandlockouts

\usepackage{cite}
\usepackage{amsmath,amssymb,amsfonts}
\usepackage{algorithmic}
\usepackage{graphicx}
\usepackage{textcomp}
\usepackage{xcolor}
\usepackage{amsmath} 
\usepackage{hyperref}

\def\BibTeX{{\rm B\kern-.05em{\sc i\kern-.025em b}\kern-.08em
    T\kern-.1667em\lower.7ex\hbox{E}\kern-.125emX}}
\begin{document}

\title{HandS3C: 3D Hand Mesh Reconstruction \\ with State Space Spatial Channel Attention \\ from RGB images
}


\author{
Zixun Jiao$^{1}$ \qquad Xihan Wang$^{1}$ \qquad Zhaoqiang Xia$^{2}$
\qquad Lianhe Shao$^{1}$ \qquad Quanli Gao$^{1,*}$
\\
$^{1}$Xi’an Polytechnic University, Xi’an, China,
$^{2}$Northwestern Polytechnical University, Xi’an, China \\
$^{*}$corresponding author
}

\maketitle

\begin{abstract}
Reconstructing the hand mesh from one single RGB image is a challenging task because hands are often occluded by other objects. Most previous works attempt to explore more additional information and adopt attention mechanisms for improving 3D reconstruction performance, while it would increase computational complexity simultaneously. To achieve a performance preserving architecture with high computational efficiency, in this work, we propose a simple but effective 3D hand mesh reconstruction network (i.e., HandS3C), which is the first time to incorporate state space model into the task of hand mesh reconstruction. In the network, we design a novel state-space spatial-channel attention module that extends the effective receptive field, extracts hand features in the spatial dimension, and enhances regional features of hands in the channel dimension. This helps to reconstruct a complete and detailed hand mesh. Extensive experiments conducted on well-known datasets facing heavy occlusions (such as FREIHAND, DEXYCB, and HO3D) demonstrate that our proposed HandS3C achieves state-of-the-art performance while maintaining minimal parameters. Code can be available in \href{https://github.com/JiaoZixun/HandS3C}{https://github.com/JiaoZixun/HandS3C}.
\end{abstract}

\vspace{-0.2cm}
\section{Introduction}
The reconstruction of 3D hand mesh from one single RGB image has a wide range of applications in many fields, such as VR/AR, robotics, and human-computer interaction and has attracted much attention due to its low cost and computational friendliness. Influenced by the rapid development of deep learning techniques, many excellent works \cite{Hasson, Boukhayma, Choutas, Hampali, Wang_HandGCAT, JoonKyu, Mengcheng_Interacting} have emerged for this task. These works focus on important issues related to hand-object occlusion and the loss of local information. Despite significant advancements, most methods still struggle to reconstruct hand poses in some scenarios with substantial occlusion or when the prior knowledge is not sufficient.


The previous methods can be broadly categorized into two categories. One is to incorporate additional prior knowledge \cite{Wang_HandGCAT, Zimmermann_2017, MS-MANO, Iqbal}. To cite few, Wang et al. \cite{Wang_HandGCAT} added keypoint coordinates as strong prior information to constrain hand pose. Zimmermann et al. \cite{Zimmermann_2017} obtained hand region features via a hand segmentation network. Xie et al. \cite{MS-MANO} combined physiological aspects of hand motion to generate kinematic-compliant hand postures. These methods can enhance pose reconstruction performance, but lack essential prior information in practical applications (e.g., VR/AR games, robotics). Another class of approaches is to learn potential associations of image features through deeper architectures, but inevitably there is a scarcity of real annotation data. Some researchers \cite{JoonKyu,Zixun_MTHI,Simple_Baseline} use Transformer architecture to obtain global information and expect to learn hand feature mapping from occluded parts. Other researchers \cite{Pavlakos_Hands} aggregated a large number of real datasets containing annotations for exploring associations between image features. However, these studies still have some limitations. First, the attention mechanism leads to excessive computational complexity. Second, an overload of global information can interfere with the effective receptive field \cite{Wenjie}.

\begin{figure*}[htbp]
  \vspace{-0.2cm}
  \centering
  \includegraphics[width=18cm]{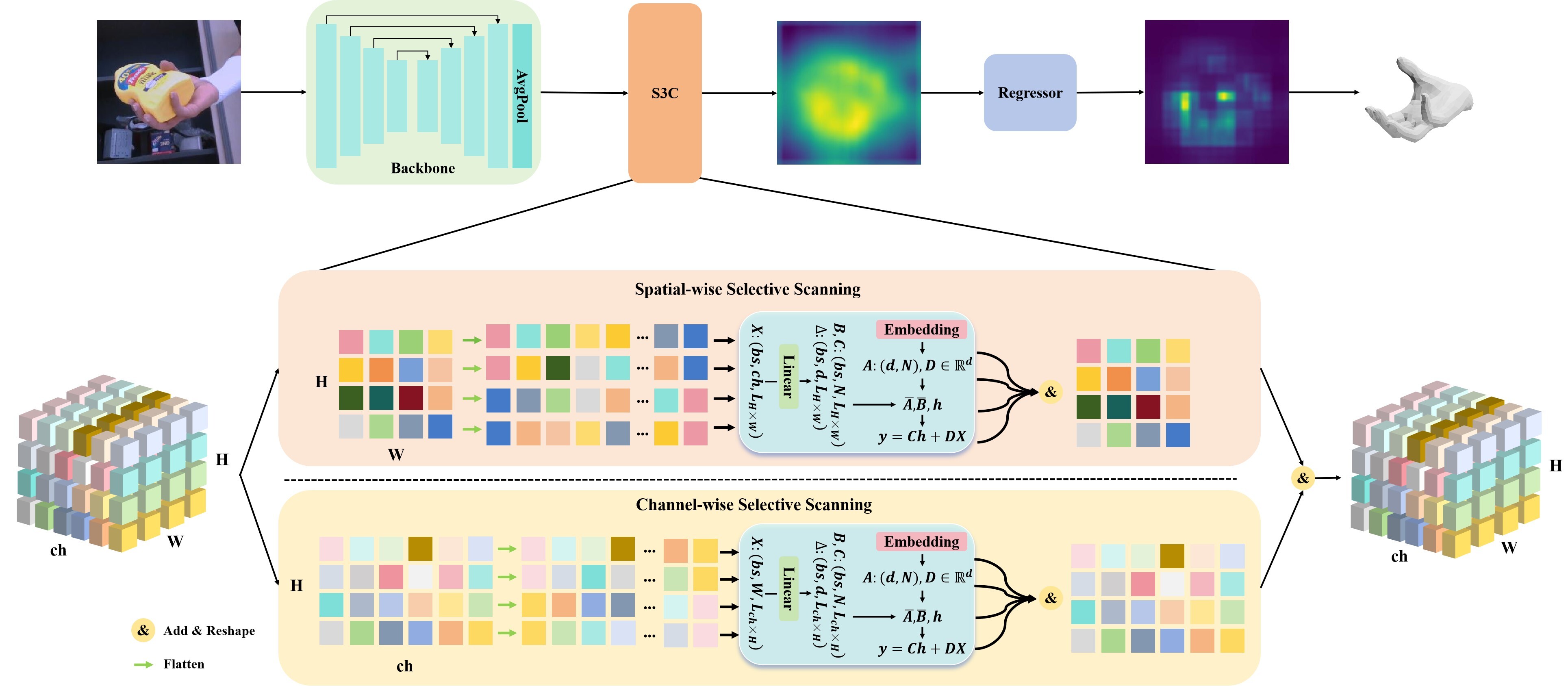}
  \caption{Overall architecture of HandS3C including Backbone, S3C and Regressor. In \textit{Backbone}, ResNet-50 based encoder-decoder is employed to extract image features and fuses multiscale features. In \textit{S3C}, image features are fed into Spatial-wise Selective Scanning and Channel-wise Selective Scanning modules, respectively. In \textit{Regressor},  heat map features are obtained using convolution operation . Further MANO parameters are obtained from the FC layer to generate the hand mesh.}
  \label{fig2}
  \vspace{-0.8cm}
\end{figure*}

Based on the above observations, we propose a simple but effective 3D hand mesh reconstruction network called as \textbf{HandS3C}. This framework incorporates the state space model \cite{VMamba, U-Mamba, VM-UNet, Mamba} into the field of hand mesh reconstruction for the first time. At the same time, combining the idea of spatial-channel feature fusion \cite{Squeeze}, \textbf{S}tate \textbf{S}pace \textbf{S}patial \textbf{C}hannel (S3C) attention module is constructed to emphasize the contribution of channel-dimensional features. This not only increases the range of effective receptive fields in the network, but also effectively avoids the square-step computational complexity of traditional attention. Specifically, our main contributions are summarized as follows:\\
\vspace{-0.5cm}
\begin{itemize}
\item We design a 3D hand mesh reconstruction network called as HandS3C, which introduces the state space model to the task of hand mesh reconstruction for the first time. This network can effectively improve hand reconstruction performance without the need for additional prior knowledge.\\
\item We propose a spatial and channel-wise parallel scanning approach to compensate for the missing channel information in planar scanning. The state-space spatial-channel attention module is constructed, which can enhance the effective receptive field range while maintaining small number of parameters.\\
\item Our method achieves state-of-the-art performance on three datasets, FREIHAND, DEXYCB and HO3D, with smaller number of parameters.
\end{itemize}

\section{Methodology}
\subsection{Overview}
The overall network architecture is shown in Fig. \ref{fig2} and consists of Backbone, S3C and Regressor components. Given a hand image $I\in\mathbb{R}^{256\times256\times3}$, the appearance information is firstly extracted by ResNet-50 based encoder-decoder and the multi-scale features are fused layer by layer to get the original feature map $F\in\mathbb{R}^{256\times32\times32}$. Then, it is fed into the S3C module and scanned in both spatial and channel dimensions to compute a 2D feature map and fused to obtain enhanced hand features $H\in\mathbb{R}^{256\times32\times32}$. Finally, the hand features $H$ are passed to the Regressor to calculate the keypoint heat map $HM\in\mathbb{R}^{21\times32\times32}$ to predict the keypoint coordinates. For the final 3D hand mesh, we use the most popular MANO parametric hand \cite {Romero_mano}. It is an articulated hand network generated from differentiable functions that takes as input the parameters that control the shape $\beta\in\mathbb{R}^{10}$ and pose $\theta\in\mathbb{R}^{48}$ of the hand.

\subsection{Backbone}
\vspace{-0.1cm}
In the original feature extraction part, we use ResNet-50 to extract image information and fuse multi-scale features layer by layer in an U-shape encoder-decoder structure. Specifically, when the input image size is $256\times256\times3$, the size of the first layer of features after ResNet-50 becomes $128\times128\times128$. In the subsequent layers, each layer of image features $F_i\in\frac{H_i}{2^i}\times\frac{W_i}{2^i}\times C_i,i\in(1,2,3,4),C_i\in(256,512,1024,2048)$ is obtained by downsampling. Then, upsampling is performed and the image features $F_{i}$ are incorporated layer by layer to get the last layer of feature extraction with size of $256\times64\times64$. Finally, the feature mapping map $F\in\mathbb{R}^{256\times32\times32}$ is obtained by fusing the feature information through average pooling.

\subsection{State Space Spatial Channel (S3C) Attention Module}
\paragraph{Spatial-wise Selective Scanning} Unlike textual information, image information contains 2D spatial information such as local texture and global structure, and it is difficult to obtain the contextual relationship of the local sense field of an image by scanning with a simple S6 module. The 1-D convolution is extended to 2-D convolution in S4ND \cite{S4ND}, and this approach leads to the loss of the ability to dynamically update weights based on context. In VMamba \cite{VMamba}, rows and columns are unfolded and scanned along in four different directions: top-left to bottom-right, bottom-right to top-left, top-right to bottom-left, and bottom-left to top-right, by means of a cross-scanning module. This allows any pixel to contain contextual information from all four directions and can enhance the size of the sensory field in the $H\times W$ plane. Sequences in each of the four directions are computed in the Spatial-wise Selective Scanning to obtain $B,C\in\mathbb{R}^{bs\times N\times L_{H\times W}}$, and $\Delta\in\mathbb{R}^{bs\times d\times L_{H\times W}}$, which in turn computes the feature map. Finally, the four sequences are merged to obtain a new sequence with the same size as the input sequence.

\paragraph{Channel-wise Selective Scanning} Although spatial features such as local texture and global structure intrinsic to 2D image features are already available through Liu et al \cite{VMamba}. However, in deep features, channels contain richer and diverse contextual information, which is mostly ignored by these methods. Therefore, we design a spatial and channel-based parallel scanning approach to compensate for the missing channel information in planar scanning, as shown in Fig. \ref{fig2}. We perform scan expansion and scan fusion on the $ch\times H$ plane. First, the input features are expanded on the $ch\times H$ plane and scanned from four directions to obtain the sequence $X_1,X_2,X_3,X_4$. Then, in Channel-wise Selective Scanning, $B,C\in\mathbb{R}^{bs\times N\times L_{ch\times H}}$, and $\Delta\in\mathbb{R}^{bs\times d\times L_{ch\times H}}$ is calculated by Eq. \ref{eq:matrxb1}-\ref{eq:matrxb3} as an indication of the correlation between the feature channels. Finally, the sequences are merged and reconstructed into feature maps with the same size for images.

Image features are spatially selected by activating the mapping of each pixel in the four directions to obtain local texture and structure information. In channel selection, the mapping relationship between different channels can be obtained to enhance the representation of similar channels in the implicit space in order to construct a wide range of global feature mappings and to enhance the scope of the sensory field. The novel state space spatial channel attention module is able to maintain a larger and deeper receptive field and global feature mapping capability with a reduced number of parameters and linear complexity. This is consistent with our motivation to expect that enhancing the effective receptive field can better capture hand-object information associations from image features.
\vspace{-0.2cm}
\begin{equation}
\label{eq:matrxb1}
h^{'}(t)=Ah(t)+Bx(t) , \; y(t)=Ch(t)+Dx(t)
\end{equation}
\vspace{-0.3cm}
\begin{equation}
\label{eq:matrxb2}
\begin{aligned}
h_{i}=&\overline{A}h_{i-1}+\overline{B}x_{i} , y_i=Ch_i+Dx_i , \\
\overline{A}=&e^{\Delta A} ,
\overline{B}=(e^{\Delta A}-I)A^{-1}B ,
\overline{C}=C 
\end{aligned}
\end{equation}
\vspace{-0.3cm}
\begin{equation}
\label{eq:matrxb3}
B=(e^{\Delta A}-I)A^{-1}B\approx(\Delta A)(\Delta A)^{-1}\Delta B=\Delta B
\end{equation}

\subsection{Regressor}
In Regressor, we input the hand attention features obtained from S3C into it. First, the 2D joint heat map $HM$ is computed using a convolution operation. Then, a cascade of hand attention features and a 2D heatmap $HM$ is input into four residual blocks. Finally, the pose parameter $\theta\in\mathbb{R}^{48}$ and the shape parameter $\mathfrak{\beta}\in\mathbb{R}^{10}$ of the MANO are predicted. Detailed 3D hand joint coordinates $J\in\mathbb{R}^{21\times3}$ and 3D hand mesh $V\in\mathbb{R}^{778\times3}$ are obtained from the MANO hand model generated.\\

\vspace{-0.3cm}
To train our proposed HandS3C model for hand mesh reconstruction, we constrain the model by calculating L2 loss.
\begin{equation}
\begin{aligned}
\mathrm{Loss}=\alpha_{\theta}\|\theta^{\prime}-&\theta\|+\alpha_{\beta}\|\beta^{\prime}-\beta\|+\alpha_{2d}\|J_{2d}^{\prime}-J_{2d}\| 
\\ +&\alpha_{3d}\|J_{3d}^{\prime}-J_{3d}\|+\alpha_{V}\|V^{\prime}-V\|
\end{aligned}
\end{equation}
where $*^{\prime}$ denotes the true value, $\text{*}$ denotes the predicted value, and $\alpha_{\theta}=10,\alpha_{\beta}=0.1,\alpha_{2d}=1\mathrm{e}2,\alpha_{3d}=1\mathrm{e}4,\alpha_{V}=1\mathrm{e}4$. the optimal solution is obtained by minimizing this objective function for training.

\begin{table}[htbp]
  \setlength{\abovecaptionskip}{-0.5cm}
  \caption{COMPARISON WITH STATE-OF-THE-ART METHODS ON FREIHAND.}
  \label{table1}
  \begin{center}
  \resizebox{1.0\linewidth}{!}{
  \begin{tabular}{c|ccccc}
    Method& Params↓& PA-& PA-& F@5↑& F@15↑\\
        & & MPJPE↓ & MPVPE↓ & & \\
    \hline
    Hasson et al. \cite{Hasson} & 33.5M  & 13.3     & 13.3 & 0.429  & 0.907\\
    ExPose \cite{Choutas}     & -- & 11.8      & 12.2     & 0.484 & 0.918\\
    Kulon et al. \cite{Weakly_Kulon}     & 519.0M       & 8.4  & 8.6     & 0.614       & 0.966\\
    I2L-MeshNet \cite{I2L-MeshNet}     & 136.8M       & 7.4  & 7.6     &0.681       & 0.973\\
    Pose2Mesh \cite{Pose2Mesh}     & 75.0M       & 7.4  & 7.6    & 0.683       & 0.973\\
    Tang et al. \cite{Towards}     & 156.5M       & 6.7  & 6.7     & 0.724       & 0.981\\
    Junhyeong et al. \cite{Cross-Attention}     & 153.0M       & \textbf{6.5}  & --     & --       & 0.982\\
    AHRNET \cite{AHRNET}    & --    & 7.2   & --    & --    & 0.978\\
    \hline
    Baseline & &8.0  & 7.4     & 0.669 & 0.977 \\
    Baseline+Self-Attention  &   & 7.7 & 7.0  & 0.688  & 0.981 \\
    HandS3C (Ours)     & \textbf{31.8M}       & 7.0  & \textbf{6.3}     & \textbf{0.727}       & \textbf{0.989}\\
    \hline
  \end{tabular}
  }
  \end{center}
\end{table}

\begin{table}[htbp]
  \setlength{\abovecaptionskip}{-0.1cm}
  \vspace{-0.3cm}
  \caption{COMPARISON WITH STATE-OF-THE-ART METHODS ON DEXYCB.}
  \label{table2}
  \centering
  \begin{tabular}{c|c c c}
    Method     & Params↓     & MPJPE↓  & PA-MPJPE↓ \\
    \hline
    Spurr et al. \cite{Spurr_Weakly} & --  & 17.34     & 6.83 \\
    METRO \cite{End-to-End}     & 229.5M & 15.24      & 6.99  \\
    Liu et al. \cite{Semi-Supervised_2021}     & 34.9M       & 15.28  & 6.58 \\
    HandOccNet \cite{JoonKyu}     & 39.3M       & 14.04  & 5.80  \\
    H2ONet \cite{H2ONet}   & - & 14.00 & 5.70 \\
    Feng et al. \cite{Feng} &- & 13.20 & 5.60 \\
    HandGCAT \cite{Wang_HandGCAT}     & 75.8M       & 13.76  & 5.60 \\
    MS-MANO \cite{MS-MANO}     & --       & 12.92  & -- \\
    \hline
    Baseline &  & 14.07   & 5.78 \\
    Baseline+Self-Attention     &   & 14.40 & 5.81 \\
    HandS3C (Ours)     & \textbf{31.8M}       & \textbf{12.81}  & \textbf{5.52}\\
    \hline
  \end{tabular}
  \vspace{-0.3cm}
\end{table}

\begin{table*}[htbp]
  \setlength{\abovecaptionskip}{0cm}
  \vspace{-1cm}
  \caption{COMPARISON WITH STATE-OF-THE-ART METHODS ON HO3D.}
  \label{table3}
  \centering
  \begin{tabular}{c|c c c c c c c c}
    Method     & Params↓     & PA-MPJPE↓ & PA-MPJPE AUC↑ & PA-MPVPE↓  & PA-MPVPE AUC↑   & F@5↑     & F@15↑\\
    \hline
    I2L-MeshNet \cite{I2L-MeshNet} & 136.8M  & 11.2  & 0.775  & 13.9  & 0.722  & 0.409  & 0.932\\
    Hasson et al. \cite{Leveraging_Hasson}  & \textbf{13.2M}  & 11.0  & 0.780  & 11.2  & 0.777  & 0.464  & 0.939\\
    Hampali et al. \cite{HOnnotate}  & --  & 10.7  & 0.788  & 10.6  & 0.790  & 0.506  & 0.942\\
    METRO \cite{End-to-End}  & 229.5M  & 10.4  & 0.792  & 11.1  & 0.779  & 0.484  & 0.946\\
    Liu et al. \cite{Semi-Supervised_2021}  & 34.9M  & 9.9  & 0.803  & 9.5  & 0.810  & \textbf{0.528}  & 0.956\\
    I2UV-HandNet \cite{I2UV-HandNet}  & --  & 9.9  & 0.804  & 10.1  & 0.799  & 0.500  & 0.943\\
    ArtiBoost \cite{ArtiBoost}  & 25.3M  & 11.4  & 0.773  & 10.9  & 0.782  & 0.488  & 0.944\\
    Keypoint \cite{Hampali}  & 56.5M  & 10.8  & 0.786  & --  & --  & --  & --\\
    \hline
    Baseline    & & 9.8 & 0.798  & 9.7 & 0.801    & 0.516 & 0.949 \\
    Baseline+Self-Attention(4.6M)     & & 10.1 & 0.798 & 10.1 & 0.798  & 0.502 & 0.947 \\
    Baseline+SSA(0.57M)        & & 9.8 &0.803 & 9.8 &0.804  & 0.521 & 0.952\\
    Baseline+serial-S3C(0.58M)     & & 9.9   &--     & 9.8 &--   & 0.516 & 0.952\\
    Baseline+S3C(0.58M) (\textbf{Ours})  & 31.8M  & \textbf{9.5}  & \textbf{0.808}  & \textbf{9.5}  & \textbf{0.810}  & 0.518  & \textbf{0.957}\\
    \hline
  \end{tabular}
  \vspace{-0.5cm}
\end{table*}

\vspace{-0.3cm}
\section{Experiments}
\subsection{Implementation Details}
We train on three publicly available datasets, FREIHAND\cite{zimmermann2019freihand}, DEXYCB\cite{chao2021dexycb}, and HO3D\cite{HOnnotate}, and the model was trained by annealing every 10th time starting from an initial learning rate of 10$^{-4}$ to train the model with batch size of 32. Our training and testing were performed on a server equipped with a GPU RTX A5000 (24GB). All other details will be provided in our code. In the FREIHAND, we report PA-MPJPE and PA-MPVPE and F-scores in $mm$. For DEXYCB, we report MPJPE and PA-MPJPE in $mm$. For HO3D, we report PA-MPJPE/MPVPE, AUC and F scores returned from the official evaluation server.

\begin{figure}[htbp]
  \centering
  \includegraphics[width=7.5cm]{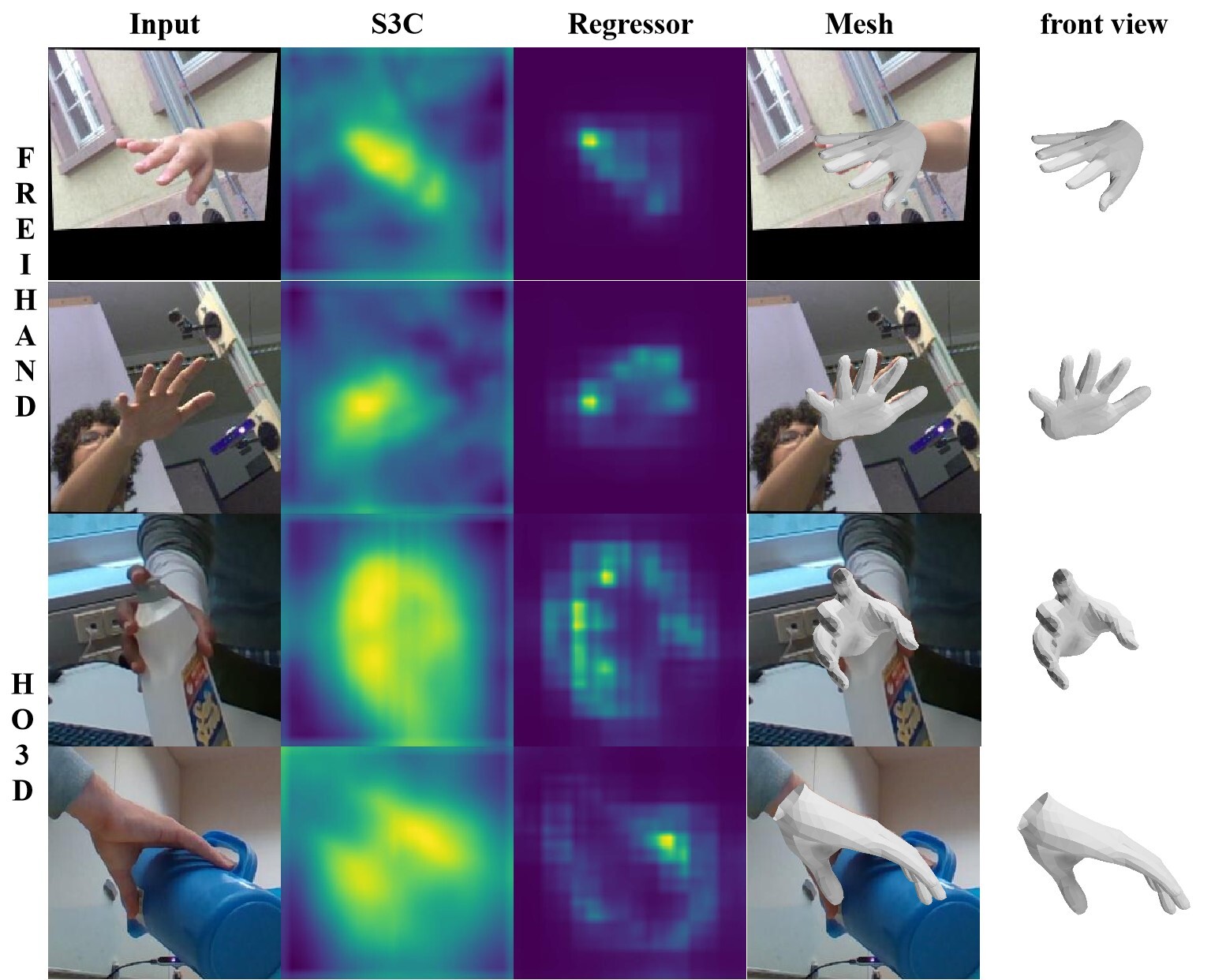}
  \caption{Qualitative results for FREIHAND and HO3D. From left to right are the input image containing severe occlusions and angular distortions, the attention mapping map obtained by S3C, the heat map of keypoints in Regressor, and the hand mesh.}
  \label{fig3}
\end{figure}

\subsection{Comparisons with the State-of-the-art Methods}
\begin{figure}[htbp]
  \vspace{-0.2cm}
  \centering
  \includegraphics[width=8cm]{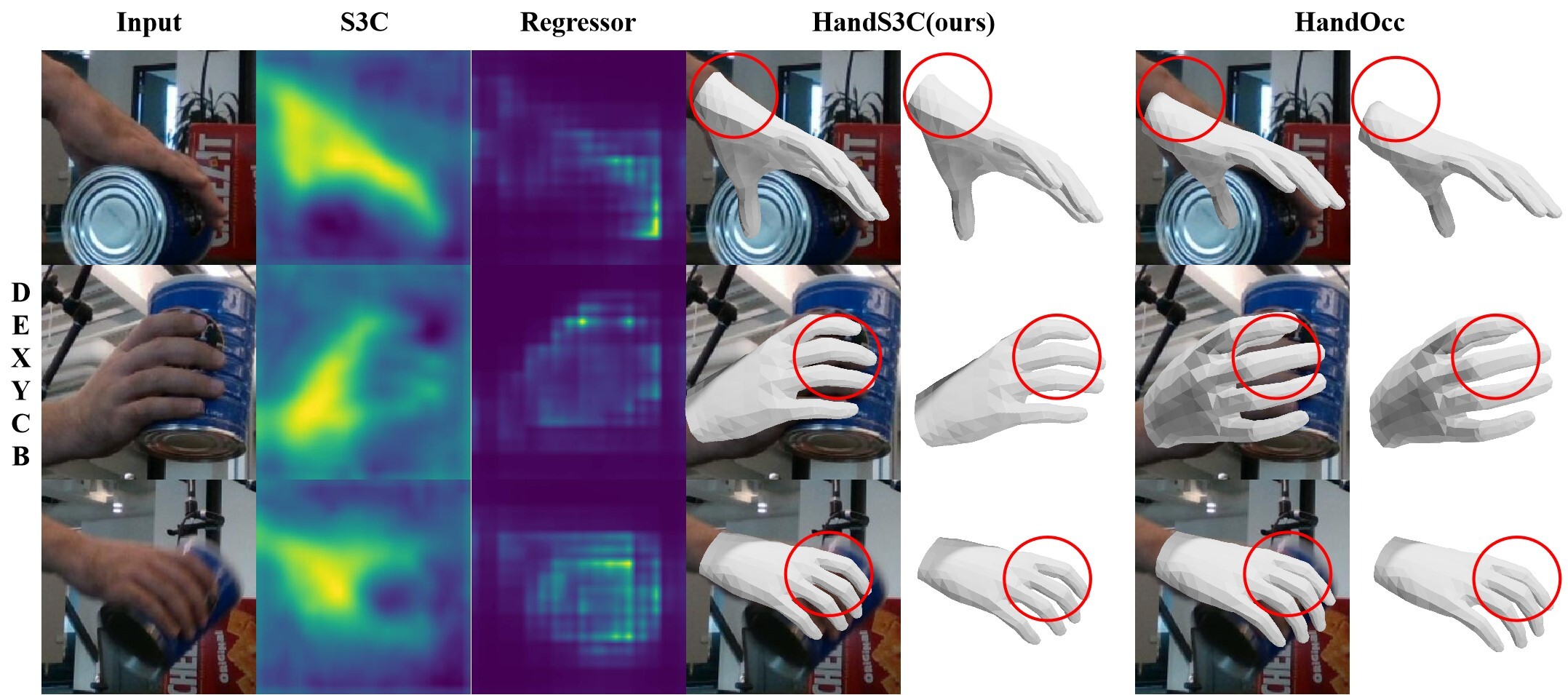}
  \caption{Qualitative results for DEXYCB. From left to right are the input image containing severe occlusions and angular distortions, the attention mapping map obtained by S3C, the heat map of keypoints in Regressor, the hand mesh and HandOCC \cite{JoonKyu} hand mesh results.}
  \label{fig4}
  \vspace{-0.2cm}
\end{figure}

We compare our proposed method with several recent state-of-the-art methods on three datasets as shown in Tables \ref{table1}-\ref{table3}. As can be seen in table \ref{table1}, our method only slightly underperforms the other methods\cite{Cross-Attention} in the PA-MPJPE metrics, while remaining ahead in the other metrics, and it is important to note that the amount of parameters in our method has been reduced by a factor of almost 5\cite{Cross-Attention}. On the HO3D dataset, our method reduces the number of parameters by about 40\% compared to recent methods \cite{Hampali}. Meanwhile, when the number of parameters is similar to the method of Liu et al. \cite{Semi-Supervised_2021}, all other metrics are greatly improved, and only the F-score metric is slightly reduced. To verify the effectiveness of our proposed method in the case of large-scale occlusion, the metrics on the DEXYCB dataset are shown in Table \ref{table2}, and all the metrics outperform the recent state-of-the-art models \cite{Wang_HandGCAT, MS-MANO, H2ONet, Feng}. In conclusion, the experimental results on all three datasets show that our proposed method can still be effective in estimating hand postures when a large number of occlusions are included. Moreover, the state-of-the-art performance can be maintained while reducing the number of parameters, as shown in fig.\ref{fig3} - \ref{fig4}.

\subsection{Ablation Study}
In order to verify the effectiveness of the HandS3C method, we further analyzed different attention modules. On the FreiHand and DEXYCB datasets, we conducted three experiments: baseline, adding a self-attention module, and adding an S3C module. On the HO3D dataset, we supplemented these three ablation experiments with experiments on the traditional state-space attention module (SSA) and the serial state-space spatial channel attention module (serial-S3C). With these ablation experiments, we demonstrate that the state-space model with added channel information can improve the ability to extract occluded partial hand features while extending the effective sensory field enhancement, and verify the effectiveness of the parallel structure. The results are summarized in Tables \ref{table1}-\ref{table3}.

From Tables \ref{table1} and \ref{table3}, it can be seen that on smaller scale datasets, the self-attention mechanism is difficult to learn the global features of the occluded part, but instead introduces more noise information to cause interference. The S3C module, on the other hand, expands the range of the sensory field through the spatial plane selection mechanism, and then enhances the hand region features directionally through the selection of the channel plane, which in turn avoids the interference of global noise. From Table \ref{table3}, it can be found that compared with the original SSM, the number of parameters of the model is only slightly increased by 0.01M after the introduction of the state selection of the channel plane, but a significant enhancement of nearly 2\% is realized in the two performance metrics of PA-MPJPE and PA-MPVPE. This is due to the fact that our approach emphasizes the importance of channel information, which in combination with spatial information can expand the effective sensory field and thus extract the features of the region of focus.

The study of Deng et al. successfully realized image restoration through serial structure \cite{CU-Mamba}. However, since SSM resembles the linear computation of recurrent neural networks (RNN). Serial scanning in space and channel add extra computation time. Meanwhile, compared to the parallel structure, the serial structure does not balance the contributions of spatial and channel dimensions well, as shown in the ablation experiments in Table \ref{table3}.
\section{Conclusion}
In order to effectively alleviate the challenges of the hand mesh reconstruction task under occlusion conditions, we firstly introduce the state space model into the hand mesh reconstruction task and propose the HandS3C method. The method enlarges the effective receptive field by performing parallel cross-scanning of the spatial plane and the channel plane. Thus, hand features under occlusion conditions are acquired while keeping fewer parameters. Extensive experimental results show that this approach can accurately predict hand pose under heavy occlusion. The ablation study also validates the effectiveness of the space state model in comparison to the attention-based model.
\section{Acknowledgment}
This work was supported by the National Natural Science Foundation of the China, Shaanxi Provincial Key Industry Innovation Chain Program, 2024 public digital cultural service research project of the national public cultural development center of the Ministry of culture and tourism, and the scientific research program of Shaanxi Provincial Department of Education [grant numbers 62072362 and 12101479, 2020ZDLGY07-05, GGSZWHFW2024-017, 23jp060].

\end{document}